%% file: main.tex
\documentclass[11pt]{article}
\usepackage{amscd,amssymb,stmaryrd}

\usepackage{graphicx}
\usepackage{subcaption}
\usepackage[utf8]{inputenc}
\usepackage[export]{adjustbox}
\usepackage{wrapfig}

\usepackage{amsthm}
\usepackage{bbm}

\usepackage{color}
\usepackage{mathtools}

\usepackage{hyperref}
\usepackage[english]{babel}
\usepackage{booktabs}

\usepackage{float}
\usepackage[linesnumbered,ruled,vlined, ruled,vlined]{algorithm2e}

\newtheorem{theorem}{Theorem}[section]

\theoremstyle{definition}

\usepackage{tikz}
\usepackage{pifont}%

\usepackage{diagbox}
\usetikzlibrary{decorations.pathreplacing,calligraphy}
\usepackage{pgfplots}
\pgfplotsset{compat=1.11}
\usepgfplotslibrary{groupplots}

\usetikzlibrary{shapes.misc}

\tikzset{cross/.style={cross out, draw=black, minimum size=2*(#1-\pgflinewidth), inner sep=0pt, outer sep=0pt},
cross/.default={1pt}}

\usepackage[linesnumbered,ruled,vlined, ruled,vlined]{algorithm2e}
\usepackage{amsmath}
\usepackage{lineno}
\usepackage{subcaption}
\usepackage{tikz}
\usepackage{xcolor}
\usepgfplotslibrary{fillbetween}
\usepackage{float}
\usepackage{pgfplots}
\pgfplotsset{compat=newest} 

\title{Conformalized-DeepONet: A Distribution-Free Framework for Uncertainty Quantification in Deep Operator Networks}

\author{Christian Moya \footnote{Department of Mathematics, Purdue University, West Lafayette, IN 47907, USA. (Email: cmoyacal@purdue.edu)}, Amirhossein Mollaali \footnote{School of Mechanical Engineering, Purdue University, West Lafayette, IN 47907, USA. (Email: amollaal@purdue.edu)}, Zecheng Zhang\footnote{Corresponding author. Department of Mathematics, Florida State University, Tallahassee, FL 32304, USA. (Email: zecheng.zhang.math@gmail.com)}, Lu Lu\footnote{Department of Statistics and Data Science, Yale University, New Haven, CT 06511, USA. (Email: lu.lu@yale.edu)}, Guang Lin \footnote{Department of Mathematics and Mechanical Engineering, Purdue University, West Lafayette, IN 47907, USA. (Email: guanglin@purdue.edu)}
}

\begin{document}
\maketitle

\begin{abstract}
In this paper, we adopt conformal prediction, a distribution-free uncertainty quantification (UQ) framework, to obtain confidence prediction intervals with coverage guarantees for Deep Operator Network (DeepONet) regression. Initially, we enhance the uncertainty quantification frameworks (B-DeepONet and Prob-DeepONet) previously proposed by the authors by using split conformal prediction. By combining conformal prediction with our Prob- and B-DeepONets, we effectively quantify uncertainty by generating rigorous confidence intervals for DeepONet prediction. Additionally, we design a novel Quantile-DeepONet that allows for a more natural use of split conformal prediction. We refer to this distribution-free effective uncertainty quantification framework as split conformal Quantile-DeepONet regression. Finally, we demonstrate the effectiveness of the proposed methods using various ordinary, partial differential equation numerical examples, and multi-fidelity learning.
\end{abstract}
\section{Introduction} \label{sec:introduction}
Recent advancements in neural network technology have solidified its standing as a reliable and efficient tool for function approximation~\cite{barron1993universal, rahimi2008weighted}. This is particularly evident in scientific computing, where neural networks have demonstrated their power in approximating solutions to Partial Differential Equations (PDEs) or Ordinary Differential Equations (ODEs) \cite{raissi2019physics, lu2021deepxde, karniadakis2021physics, leung2022nh, rathore2024challenges}. A prevalent challenge in scientific computing involves approximating operators that map one function to another. Seminal research endeavors, such as those by Chen and Chen \cite{chen1995universal, chen1993approximations}, have explored the utilization of neural networks to approximate such operators, commonly referred to as neural operators.

In recent years, there has been a surge of interest in neural operators, with notable contributions from various scholars \cite{lu2021learning, lu2022comprehensive, lu2022multifidelity, zhu2023fourier, li2020fourier, li2020neural, wen2022u, li2022fourier}. Among these, two prominent approaches have emerged as particularly successful: Deep Operator Neural Networks (DeepONet) \cite{lu2021learning, lu2022comprehensive, jin2022mionet, lu2022multifidelity, lin2023learning, zhu2023reliable} and Fourier Neural Operator (FNO) \cite{lin2023b, zhang2023d2no, li2020fourier, li2020neural, wen2022u, li2022fourier, pathak2022fourcastnet}.

Compared to FNO, DeepONet represents a mesh-free neural operator, implying that the output functions do not require discretization. This characteristic enhances the flexibility of DeepONet in both training and testing phases \cite{lin2021operator, zhang2022belnet, zhang2023discretization, lu2022comprehensive, moya2023deeponet, hayford2024speeding, sun2023deepgraphonet, zhang2023d2no}. Furthermore, recent advancements in BelNet by \cite{zhang2022belnet, zhang2023discretization, zhang2023d2no} have extended DeepONet's capabilities, making it invariant to input function discretization. This means that input functions no longer need to conform to a shared discretization, thereby enhancing the versatility of DeepONet even further.

DeepONet and its extensions have been successfully applied to a diverse array of real-world applications. These include weather forecasting \cite{pathak2022fourcastnet}, sub-surface structure detection \cite{zhu2023fourier}, electrical propagation on the left ventricle \cite{yin2024dimon}, geological carbon sequestration \cite{jiang2023fourier}, disk-planet interactions in protoplanetary disks \cite{mao2023ppdonet}, hypersonic systems \cite{mao2021deepm, di2023neural}, power systems \cite{moya2023approximating}, and optimization \cite{sahin2024deep}. 

The need for Uncertainty Quantification (UQ) in scientific machine learning~\cite{psaros2023uncertainty,zou2024neuraluq} often arises from several factors. These include uncertainties in data due to measurement errors and numerical algorithm errors, uncertainties in model forms due to network architectures and physical models of varying fidelities, and uncertainties in parameters due to network training and system properties at different scales. As a result, researchers are tasked with providing confidence intervals for predicted outputs. A good interval should meet two fundamental criteria. Firstly, it should have a significant coverage rate, encompassing precise solutions or dynamics to the greatest extent possible. Secondly, the interval should be minimized in size.

Numerous works~\cite{lin2023b,moya2023deeponet,psaros2023uncertainty,zou2024neuraluq,yang2022scalable,garg2023vb,guo2023ib,sahin2024deep,zhang2023homogenization,moya2023bayesian} have established various uncertainty quantification (UQ) frameworks for DeepONet. These frameworks aim not only for accurate prediction but also for quantifying uncertainty. For example, a recent study~\cite{lin2023b} from the authors addresses this uncertainty issue in operator learning by framing the training process within a Bayesian framework. The study uses Langevin diffusion-based sampling methods \cite{welling2011bayesian, dalalyan2017further, dalalyan2019user, raginsky2017non, zheng2024accelerating} to generate ensembles, which aids in characterizing uncertainty in DeepONet predictions. To lessen the use of ensembles and increase efficiency, we introduced a probabilistic framework for UQ in DeepONets in~\cite{moya2023deeponet}, which provides an input-dependent standard deviation as a heuristic measure of uncertainty. However, despite the promising results from these methods, constructing rigorous confidence intervals for DeepONet predictions remains an unsolved challenge, which we aim to solve using conformal prediction.

Conformal prediction \cite{angelopoulos2021gentle,vovk1999machine,vovk2005algorithmic} is an alternative method for constructing confidence intervals that provide nonasymptotic, distribution-free coverage guarantees. Unlike the Bayesian Framework, this approach does not require prior distribution knowledge and avoids relying on strong assumptions of large-sample asymptotic approximations, which may be difficult to justify in practical scenarios. These characteristics make conformal prediction a valuable tool for addressing scientific machine learning problems, especially those arising from engineering applications with real observed data and a lack of prior knowledge.

There are two types of conformal prediction methods: full conformal prediction \cite{vovk2005algorithmic,shafer2008tutorial} and split conformal prediction~\cite{vovk2005algorithmic,papadopoulos2002inductive}. In this paper, we focus on split conformal prediction, which operates with finite samples and enables building adaptive and validated confidence intervals. These properties have allowed split conformal prediction to become a popular approach for machine learning and uncertainty quantification tasks, including Language Modeling~\cite{quach2023conformal}, Graph Neural Networks~\cite{huang2024uncertainty}, time series~\cite{angelopoulos2024conformal}, and quantile regression~\cite{romano2019conformalized}. However, despite its potential, split conformal prediction remains underutilized in the scientific machine learning community, particularly in the context of Deep Operator Networks.

The \textit{goal} of this paper is to address the challenge of uncertainty quantification (UQ) in operator learning. It aims to showcase the implementation of split conformal prediction for constructing confidence intervals and improving existing UQ algorithms tailored for DeepONet regression, with coverage guarantees.

To achieve this goal, we make the following contributions:
\begin{enumerate}
\item We design a split conformal prediction algorithm (Section~\ref{subsec:locally-adaptive}) that enhances the previously proposed uncertainty quantification frameworks, Bayesian DeepONet (B-DeepONet) and Probabilistic DeepONet (Prob-DeepONet). This novel algorithm allows B-DeepONet and Prob-DeepONet to generate confidence intervals with coverage guarantees.
\item We then propose a new extension of DeepONet called Quantile-DeepONet (Section~\ref{subsec:CQR}), which we use in a novel Conformal Quantile-DeepONet regression algorithm to construct rigorous confidence intervals.
\item Finally, we demonstrate the effectiveness of our proposed conformalized-DeepONets by conducting multiple numerical experiments on ordinary differential equations, partial differential equations, and multi-fidelity settings.
\end{enumerate}
The remaining sections of this paper are organized as follows. Section~\ref{sec:Background-Information} provides a review of the concepts of DeepONet, B-DeepONet, and Prob-DeepONet. In Section~\ref{subsec:locally-adaptive}, we describe how to use split conformal prediction on B-DeepONet and Prob-DeepONet to build confidence intervals with coverage guarantees. Section~\ref{subsec:CQR} presents the design of the Quantile DeepONet, which serves as the main component of the conformal Quantile-DeepONet regression algorithm for constructing rigorous confidence intervals. We present numerical experiments that validate the performance of all conformalized DeepONets in Section~\ref{sec:numerical-experiments}. Finally, Section~\ref{sec:discussion} discusses our results and future work, and Section~\ref{sec:conclusion} concludes the paper.

\section{Background Information} \label{sec:Background-Information}
This section reviews the Deep Operator Network (DeepONet) framework~\cite{lu2021learning}. It also provides a summary of two uncertainty quantification (UQ) frameworks for DeepONet: Bayesian DeepONet (B-DeepONet~\cite{lin2023b}) and probabilistic DeepONet (Prob-DeepONet~\cite{moya2023deeponet}).
\subsection{Deep Operator Network (DeepONet)} \label{subsec:DeepONet}
The Deep Operator Network~(DeepONet)  \cite{lu2021learning, lu2022comprehensive, lu2022multifidelity, yin2024dimon} is an operator learning framework that aims to approximate the nonlinear operator $G$. The operator $G$ maps a function space $U$ with domain $K_1$ to a function space $V$ with domain $K_2$. DeepONet has been theoretically supported by universal approximation theorems \cite{chen1995universal, zhang2023discretization, jin2022mionet, deng2022approximation}. These theorems enable the approximation of the operator $G$ for a given $u \in U$ and $x \in K_2$ using the following linear trainable representation:
\begin{align*}
G(u)(x) \approx G_\theta(\hat{u})(x) = \sum_{k = 1}^{K} b_k(\hat{u})\tau_k(x),
\end{align*}
Here, $\theta$ is the vector of trainable parameters, while $b_k$ and $\tau_k$ are $K$ trainable coefficients and basis functions, respectively.

\begin{figure}[t!]
\centering
\scalebox{.78}{\input{deepo.tex}}
\caption{Stacked version DeepONet $G_{\theta}$. $\bigotimes$ denotes the inner product in $\mathbb{R}^K$.}
\label{fig_don_structure}
\end{figure}
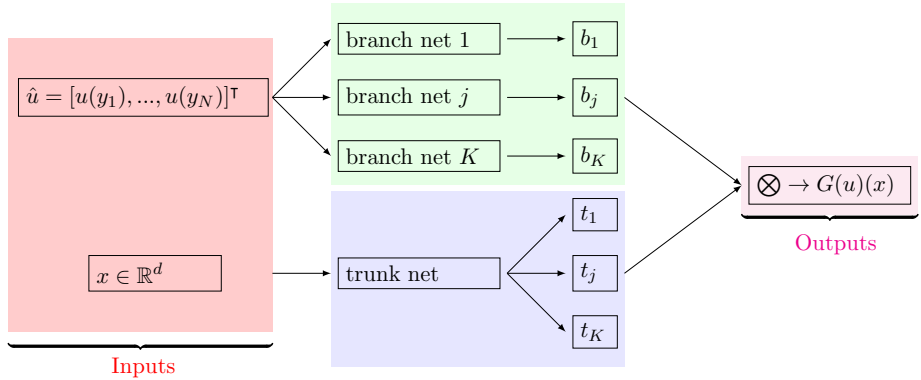

DeepONet consists of two sub-networks: the branch network and the trunk network (see Figure~\ref{fig_don_structure}). The \textit{branch} network maps the input function $u \in U$, which is discretized using $m$ sensors and denoted as~$\hat{u}$, to a vector of $K$ trainable coefficients $b(\hat{u}) \in \mathbb{R}^K$. It is worth noting that various variants of DeepONet have been proposed to handle input functions with different discretizations. For instance, the new enhanced structures proposed in \cite{zhang2022belnet, zhang2023discretization} and the training algorithms proposed in \cite{zhang2023d2no}.

On the other hand, the \textit{trunk} network maps a location $x \in K_2$, which belongs to the output function domain $K_2 \subset \mathbb{R}^d$, to a vector of $K$ trainable basis functions. This feature eliminates the need for discretization of the output function $G(u) \in V$ in DeepONet and its variants, making them mesh-free PDE approximate solvers.
To train DeepONet, one could optimize,
\begin{align*}
    \mathcal{L}(\theta) = \frac{1}{N}\sum_{i = 1}^N\sum_{j = 1}^{N_j} \left |G^{(i)} - G_\theta \left(\hat{u}^{(i)}\right)\left(x^{(i)_j}\right) \right |^2.
\end{align*}
Without loss of generality and for better illustration of the algorithms, we assume $N_j = 1$, or $x^{(i)}\in K_2$ is the only evaluation point for one function $u^{(i)}$. That is, we have the dataset,
using the dataset of $N$ triplets: $\mathcal{D} = \left\{\hat{u}^{(i)}, x^{(i)}, G^{(i)}\right\}_{i=1}^N$, where $G^{(i)} = G\left(\hat{u}^{(i)}\right)\left(x^{(i)}\right)$ is the target operator value.

Despite the remarkable accuracy of DeepONet on multiple applications (e.g., \cite{moya2023deeponet, pathak2022fourcastnet, zhu2023fourier, yin2024dimon}), it only produces pointwise predictions. However, in cases where small training datasets or noisy inputs are involved, these pointwise predictions may be unreliable. Therefore, it is necessary to have some measure of the DeepONet's uncertainty. This requirement led the authors of this paper to develop two uncertainty quantification frameworks for DeepONet: Bayesian and probabilistic DeepONets, which we review next.
\subsection{Bayesian Deep Operator Network~(B-DeepONet)} \label{subsec:B-DeepONet}
This section reviews the first framework for uncertainty quantification (UQ) in DeepONet, known as Bayesian DeepONet (B-DeepONet)~\cite{lin2023b}, developed by the authors of this paper. In particular, B-DeepONet is based on the Langevin theory that allows us to generate samples from an approximate posterior that converges to the Gibbs distribution proportional to $\exp(-U(\theta)/\tau)$, where $U$ is the non-convex energy function and $\tau >0 $ the temperature. 

Measuring the uncertainty linked with constrained training datasets, noisy inputs, and over-parametrization of neural networks poses a considerable challenge. This challenge is even more pronounced in deep operator regression, as it involves mapping between functional spaces. To address this challenge, we proposed the Bayesian DeepONet (B-DeepONet) in~\cite{lin2023b}. B-DeepONet allows us to construct heuristic confidence intervals for the operator that maps an input function $u \in V$ to an output function $G(u)(x) \in \mathbb{R}$ evaluated at a point $x \in K_2 \subset \mathbb{R}^d$.

In B-DeepONet, our goal was to develop a method to sample from the predictive distribution $p(G|(u,x),\mathcal{D})$ of the target operator $G(u)(x)$, given a discretized input $\hat{u} \in \mathbb{R}^m$ at any $x \in K_2$ using the training dataset $\mathcal{D}$. To reach this goal, we first assumed a factorized Gaussian likelihood function for the dataset~\cite{welling2011bayesian, li2023fast}:
    $$
    p(G|(\hat{u},x),\theta) = \mathcal{N}(G|G_\theta(\hat{u})(x), \text{diag}(\Sigma^2)) = \prod_{i=1}^N \mathcal{N}(G^{(i)}|G_\theta(\hat{u}^{(i)})(x^{(i)}), \sigma),
    $$
where the mean $G_\theta(\hat{u})(x)$ is the output of the underlying DeepONet with a vector of trainable parameters $\theta$ and $\text{diag}(\Sigma^2)$ a diagonal covariance matrix with $\Sigma^2 = (\sigma^2,\ldots,\sigma^2)$ on the diagonal~\cite{psaros2023uncertainty}, which we assume is given. 

It is important to note that the target operator value $G(u)(x) \in \mathbb{R}$ for a given input function $u \in U$ at any location $x \in K_2$ and the training dataset $\mathcal{D}$ is the random variable $(G|(u,x),\mathcal{D})$. To calculate the density of this random variable, the DeepONet parameters $\theta$ need to be integrated out, as shown below:
$$
p(G|(u,x), \mathcal{D}) = \int p(G|(u,x),\theta)p(\theta | \mathcal{D})d\theta.
$$
In the above, $p(\theta | \mathcal{D})$ is the posterior distribution of the DeepONet trainable parameters. Using Bayes' rule, we can write the posterior of DeepONet parameters as
$$
p(\theta|\mathcal{D}) \propto p(\mathcal{D}|\theta) p(\theta),
$$
where $p(\theta)$ is the \textit{prior} distribution of the parameters, and $p(\mathcal{D}|\theta)$ is the \textit{data likelihood}, that is, $p(\mathcal{D}|\theta) = \prod_{i=1}^N p(G^{(i)}|(\hat{u}^{(i)}, x^{(i)}), \theta)$. We calculate this using the DeepONet forward pass and the i.i.d. training dataset $\mathcal{D}$.

Computing the posterior distribution in closed form using Bayes' rule is intractable~\cite{psaros2023uncertainty} for deep operator networks. Thus, to generate samples from this distribution, we employ a Langevin diffusion-based sampling technique~\cite{raginsky2017non, chen2014stochastic}. This technique has been shown to produce samples that closely match the target distribution \cite{bhattacharya1978criteria, roberts1996exponential}. Specifically, the Langevin diffusion process used to generate samples is as follows:
\begin{align*}
d\theta_t = -\nabla U(\theta_t) dt+\sqrt{2\tau}dW_t,
\end{align*}
where $\theta$ is the vector of parameters, $U$ is the energy function (i.e., the log data likelihood and prior), $\tau$ the temperature, and $W_t$ is the Brownian motion. By sampling from the posterior distribution using a discretized Langevin process, we obtained an ensemble of $M$ $\theta$ samples, denoted as $\{\theta_k\}_{k=1}^M$, as described next.

To obtain the ensemble of parameters, denoted as $\{\theta_k\}_{k=1}^M$, B-DeepONet uses a variation of the diffusion process mentioned earlier. Particularly, the authors consider the parallel tempering techniques.  These methods involve two chains and facilitate particle swapping between them, aiding in escaping local traps and addressing highly non-convex problems, thereby enhancing its capability to tackle complex UQ challenges. However, employing two chains doubles the cost of energy function evaluation. Despite efforts by the authors to mitigate these costs, achieving only up to a $25\%$ reduction in the doubled cost, this prompts the exploration of alternative UQ methods for managing uncertainty in neural operators.

To construct the confidence intervals, in our paper~\cite{lin2023b}, we used the obtained $M$-ensemble $\{\theta_k\}_{k=1}^M$ of sampled parameters from the posterior distribution to fit a parametric predictive distribution, a common practice in the literature~\cite{psaros2023uncertainty}. It is important to note that this assumption limits the applicability of these confidence intervals. Specifically, we used the Gaussian distribution $\mathcal{N}(\mu(\hat{u})(x), \sigma(\hat{u})(x))$ for an arbitrary $x$ location. The parameters of this distribution were obtained as follows:
\begin{align*}
    \mu(\hat{u})(x) &= \frac{1}{M} \sum_{k=1}^M G_{\theta_k}(\hat{u})(x) \\
    \sigma^2(\hat{u})(x) &= \frac{1}{M}\sum_{k=1}^M \left(G_{\theta_k}(\hat{u})(x) - \mu(\hat{u})(x)  \right)^2.
\end{align*}
Due to the assumption of fitting the parametric predictive distribution, B-DeepONet can only provide a heuristic estimate of uncertainty. In our paper~\cite{lin2023b}, we demonstrated that constructing confidence intervals with B-DeepONet requires extensive hyperparameter optimization to achieve a high confidence level (e.g., 95\%) for a specific test dataset. However, there is no guarantee that these heuristic confidence intervals, developed using B-DeepONet, will be reliable for other test examples. Additionally, constructing these heuristic confidence intervals requires the forward pass of $M$ DeepONets, making B-DeepONet a computationally expensive method.
\subsection{Probabilistic Deep Operator Network~(Prob-DeepONet)} \label{subsec:Prob-DeepONet}
One of the main drawbacks of quantifying uncertainty using B-DeepONet is its cost. B-DeepONet requires performing a forward pass across all the DeepONets within the collected ensemble of size $M$. To address this drawback, the authors of this paper developed the Probabilistic DeepONet (Prob-DeepONet)~\cite{moya2023deeponet}, which we will review in this section.

Prob-DeepONet is a probabilistic model whose output approximates the parameters of a normal distribution:
$$G|X=(\hat{u}, x) \sim \mathcal{N}\left(\mu_{\theta_{\mu}}(\hat{u})(x), \sigma_{\theta_{\sigma}}(\hat{u})(x) \right).$$
In the above, $\mu_{\theta_{\mu}}$ represents the mean of the probabilistic model and aims to approximate $G(u)(x) \in \mathbb{R}$, which is the target operator for a given input $u \in U$ at $x \in K_2$. This approximation is achieved using the following linear trainable representation:
$$\mu_{\theta_\mu}(\hat{u})(x) = \sum_{i=1}^K b^{\mu}_k(\hat{u})\cdot \tau_k^{\mu}(x).$$
Similarly, $\sigma_{\theta_{\sigma}}$ represents the standard deviation of the probabilistic model. It aims to estimate the uncertainty for $G(u)(x)$ through the following linear trainable representation:
$$\log(\sigma_{\theta_\sigma}(\hat{u})(x)) = \sum_{i=1}^K b^{\sigma}_k(\hat{u})\cdot \tau_k^{\sigma}(x).$$
Note that in the above, for numerical stability, we train the model to produce $\log \sigma$ and then recover the standard deviation $\sigma$ using the exponential function.

Similar to the classical DeepONet, Prob-DeepONet consists of two sub-networks: the branch network and the trunk network. In the branch network, we design two components: one for the mean and the other for the standard deviation. The branch component for the mean includes shared trainable layers and a few independent layers. It maps the discretized input $\hat{u} \in \mathbb{R}^m$ to a vector of $K$ trainable coefficients $b^{\mu}(\hat{u}) \in \mathbb{R}^K$ for the mean. Similarly, the branch component for the log standard deviation shares several layers with the mean component. It also has independent layers to process the map from the discretized input $\hat{u} \in \mathbb{R}^m$ to the vector of trainable coefficients $b^{\sigma}(\hat{u}) \in \mathbb{R}^K$.

On the other hand, the trunk network also has two components: one for the mean and another for the log standard deviation. Each component consists of a set of shared layers and a few independent layers. The trunk component for the mean maps the location $x \in K_2$ within the output function domain $K_2 \subset \mathbb{R}^d$ to the vector of $K$ trainable basis functions $\tau^{\mu}(x) \in \mathbb{R}^K$. Similarly, the trunk component for the log standard deviation maps the location $x \in K_2$ to the vector of trainable basis functions $\tau^{\sigma}(x) \in \mathbb{R}^K$.

To train the Prob-DeepONet parameters $\theta = \{\theta_{\mu},\theta_{\sigma}\}$, we minimize the negative log likelihood:
$$\mathcal{L}_{\text{nll}}(\theta) = -\frac{1}{2N} \left(\sum_{i=1}^N \left( \frac{ \left(G^{(i)} - \mu_{\theta_{\mu}}(\hat{u}^{(i)})(x^{(i)}) \right)^2}{\sigma_{\theta_{\sigma}}(\hat{u}^{(i)})(x^{(i)})^2} + 2 \log\sigma_{\theta_{\sigma}}(\hat{u}^{(i)})(x^{(i)})\right) + N\log 2 \pi\right),$$
using the dataset of $N$ triplets: $\mathcal{D} = \left\{\hat{u}^{(i)}, x^{(i)},G^{(i)} \right\}_{i=1}^N$ .

Similar to B-DeepONet, Prob-DeepONet only provides a heuristic estimate of uncertainty through the standard deviation. As demonstrated in our paper, constructing confidence intervals using Prob-DeepONet requires heavy hyper-parameter optimization to achieve a high confidence level (e.g., 95\%) for a fixed test dataset. However, it is not guaranteed that these heuristic confidence intervals, developed using Prob-DeepONet and the estimated standard deviation, are reliable for other test examples. Therefore, the first objective of this paper is to enhance our previously proposed methods to transform these heuristic confidence intervals into rigorous prediction intervals. To accomplish this, we will employ split conformal prediction~\cite{papadopoulos2002inductive,angelopoulos2021gentle}, which will be described in the following section.
\section{Conformal Prediction for Deep Operator Network Regression} \label{sec:conformal-prediction}
In this section, we will first demonstrate how to use split conformal prediction~\cite{papadopoulos2002inductive,angelopoulos2021gentle} to create reliable confidence intervals using the heuristic uncertainty estimates~$\sigma(\hat{u})(x)$ obtained from a trained Prob-DeepONet and the $M$-ensemble generated by B-DeepONet. Next, we will propose a novel Quantile-DeepONet extension that is better suited for developing a conformalized Quantile-DeepONet regression algorithm.
\subsection{Split Conformal Prediction for DeepONet} \label{subsec:locally-adaptive}
Within this section, we employ split conformal prediction~\cite{papadopoulos2002inductive,angelopoulos2021gentle}, recognized as the predominant conformal prediction approach, facilitating Prob- and B-DeepONet in delivering robust confidence intervals. Formally, given any DeepONet test triplet $(\hat{u}_{\text{test}}, x_{\text{test}}, G_{\text{test}})$, our goal is to construct a marginal distribution-free confidence interval $\mathcal{C}(\hat{u}_{\text{test}}, x_{\text{test}})$ that likely contains the operator target $G_{\text{test}}$. That is, given a miscoverage rate $\alpha \in (0,1)$, we aim to achieve:
\begin{align} \label{eq:coverage-prob}
    \mathbb{P}\{G_{\text{test}} \in \mathcal{C}(\hat{u}_{\text{test}}, x_{\text{test}})\} \ge 1-\alpha.
\end{align}

To achieve this goal, split conformal prediction (see Algorithm~\ref{alg:locally-adaptive-conformal}) proceeds as follows.
\begin{algorithm}[h]
\DontPrintSemicolon
\caption{Split Conformal Prediction to Enhance B- and Prob-DeepONet}
\label{alg:locally-adaptive-conformal}
\textbf{Input:} Calibration data: $\{\hat{u}^{(i)}, x^{(i)},G^{(i)}\}_{i=1}^{n}$ \textcolor{black}{where $G^{(i)} = G(u^{(i)})(x^{(i)})$ is the exact output function values}, miscoverage level: $\alpha \in (0,1)$, and mean and standard deviation models: $\mu$ and $\sigma$, obtained from the $M$-ensemble of B-DeepONet or a trained Prob-DeepONet.\;
\textbf{Process:}\;
\Indp Use $\mu$ and $\sigma$ models to compute the calibration scores: 
$$s_1 = s\left(\hat{u}^{(1)}, x^{(1)}, G^{(1)}\right), \dots, s_n = s\left(\hat{u}^{(n)}, x^{(n)}, G^{(n)}\right),$$ 
where 
$$s(\hat{u},x,G) = \frac{|G - \mu(\hat{u} )(x)|}{\sigma( \hat{u} )(x)}.$$\;
Compute $\hat{q}$ as the $\frac{\lceil (n+1) (1-\alpha) \rceil}{n}-$th quantile of the calibration scores $s_1, \dots, s_n$.\;
\Indm \textbf{Return:} Rigorous confidence intervals (i.e., that satisfy \eqref{eq:coverage-prob}) for any new test example $(\hat{u}_\text{test},x_{\text{test}})$:
$$\mathcal{C}(\hat{u}_\text{test},x_{\text{test}}) = \left[\mu(u_\text{test})(x_\text{test}) - \hat{q}\sigma(u_\text{test})(x_\text{test}),\mu(u_\text{test})(x_\text{test}) + \hat{q}\sigma(u_\text{test})(x_\text{test})\right].$$
\end{algorithm}

\textit{Trained DeepONet model.} First, we have access to trained B- or Prob-DeepONet models. These models predict the target operator $G_{\text{test}}$ by using the mean output $\mu(\hat{u}_{\text{test}})(x_{\text{test}})$ and provide an estimate of the uncertainty through the standard deviation output $\sigma(\hat{u}_{\text{test}})(x_{\text{test}})$. It is important to note that, as described in our previous works~\cite{lin2023b,moya2023deeponet}, the standard deviation $\sigma$ produces a larger value when there is increased uncertainty about the test input.

\textit{Score function.} Then, given $\mu$  and $\sigma$, we define a score function for any DeepONet data triplet $(\hat{u}, x, G)$:
\begin{align} \label{eq:score-function}
s(\hat{u},x,G) = \frac{|G - \mu(\hat{u})(x)|}{\sigma(\hat{u})(x)}.
\end{align}
Note that this score function acts as a correction factor for our DeepONet measure of uncertainty, i.e., $s(\hat{u},x,G) \cdot \sigma(\hat{u})(x) = |G - \mu(\hat{u})(x)|$.

\textit{Calibration quantile.} Using a given calibration dataset of size $n$, denoted as $\{\hat{u}^{(i)}, x^{(i)}, G^{(i)}\}_{i=1}^n
$, we compute the calibration scores $s_1 = s(\hat{u}^{(1)}, x^{(1)}, G^{(1)}), \dots, s_n = s(\hat{u}^{(n)}, x^{(n)}, G^{(n)})$. Next, we calculate $\hat{q}$ as the $\frac{\lceil (n+1) (1-\alpha) \rceil}{n}$ quantile of the calibration scores $s_1, \dots, s_n$, where $\lceil \cdot \rceil$ denotes the ceil function. It is important to note that the calibration dataset must satisfy the exchangeability property~\cite{angelopoulos2021gentle}, which is true in most DeepONet settings where the training, test, and calibration data are independently and identically distributed (i.i.d.).

\textit{Confidence intervals with coverage guarantees.} We use the quantile $\hat{q}$ to construct rigorous confidence intervals for any new DeepONet test example $(\hat{u}_\text{test},x_{\text{test}})$:
\begin{align} \label{eq:confidence-intervals}
\mathcal{C}(\hat{u}_{\text{test}},y_{\text{test}}) = \{G:s(\hat{u}_{\text{test}},y_{\text{test}},G) \le \hat{q} \} \equiv \{G:|G - \mu(\hat{u}_\text{test})(x_\text{test})| \le \hat{q}\sigma(\hat{u}_\text{test})(x_\text{test}) \}.
\end{align}
These confidence intervals hold for any distribution of the data and are rigorous in the sense that they satisfy property~\eqref{eq:coverage-prob}, which was demonstrated in the conformal coverage guarantee theorem \ref{theorem:conformal} replicated next for completeness.
\begin{theorem} \label{theorem:conformal}
    \textbf{Conformal calibration coverage guarantee theorem~\cite{vovk1999machine,angelopoulos2021gentle}.} Suppose $\{X_i,Y_i\}_{i=1}^n$ and $(X_\text{test}, Y_\text{test})$ are i.i.d. Then define $\hat{q}$ as
    $$
    \hat{q} = \inf \left\{ q: \frac{|i:s(X_i,Y_i) \le q|}{n} \ge \frac{\lceil (n+1) (1-\alpha) \rceil}{n} \right\}
    $$
    and the resulting prediction sets as
    $$
    \mathcal{C}(X_\text{test}) = \{y:s(X_\text{test},y) \le \hat{q}\}.
    $$
    Then,
    $$\mathbb{P}\{ Y_\text{test} \in \mathcal{C}(X_\text{test})\} \ge 1 - \alpha.$$
\end{theorem}
We conclude this section by noting that the above theorem holds for any score function~$s$.
\subsection{Conformal Quantile DeepONet Regression} \label{subsec:CQR}
In the previous section, we demonstrated how to construct finite-sample, distribution-free, and adaptive confidence intervals of the form $\mu(\hat{u})(x) \pm\hat{q}\sigma(\hat{u})(x)$ for operator targets $G$ using the standard deviation $\sigma$ as a measure of uncertainty for DeepONets. We will show in Section~\ref{sec:numerical-experiments} that this approach effectively quantifies uncertainty in DeepONets. However, as stated in~\cite{angelopoulos2021gentle}, there is no evidence to suggest that the heuristic measure of uncertainty $\sigma$ for B- and Prob-DeepONet is strongly correlated with the quantiles of the operator target distribution. Therefore, it is more natural to expect that a DeepONet method that estimates the quantiles for the target operator $G$ distribution will enhance the use of split conformal predictions. Let us now introduce this novel Quantile-DeepONet.

\textit{Quantile-DeepONet.} Our goal is to design a Quantile-DeepONet that can approximate the conditional quantiles $t_{\alpha/2}$ and $t_{1-\alpha/2}$, given (i) a miscoverage level $\alpha \in (0,1)$, (ii) a discretized input function $\hat{u} \in \mathbb{R}^m$, and (iii) a location within the output function domain $x \in K_2$. To provide a complete understanding, let us recall the mathematical definition of a conditional quantile function. The $\gamma$th conditional quantile is defined as follows:
$$
t_\gamma(x) = \inf \{y\in \mathbb{R}:F(y|X=x) \ge \gamma\},
$$
where 
$$
F(y|X=x) = \mathbb{P}\{Y \leq y|X=x\},
$$
is the conditional distribution of $Y$ given $X=x$. In other words, $t_\gamma$ provides us with information about the percentile of the distribution of $Y$ when conditioned on $X=x$.

Similar to Prob-DeepONet, Quantile-DeepONet approximates quantiles using the following linear trainable representations:
\begin{align*} 
t_{\gamma}(\hat{u})(x) \approx \hat{t}_{\theta_\gamma}(\hat{u})(x) = \sum_{k = 1}^K b_k^{\gamma}(\hat{u})\cdot \tau_k^{\gamma}(x), 
\end{align*}
for $\gamma \in \{\alpha/2,1-\alpha/2\}$. Therefore, Quantile-DeepONet consists of two sub-networks: the branch network and the trunk network.

The \textit{branch} network comprises two components: one for the $(\alpha/2)$th quantile and another for the $(1-\alpha/2)$th quantile. Each component consists of a set of shared trainable layers and a few independent trainable layers. Additionally, each component maps the discretized input $\hat{u} \in \mathbb{R}^m$ to a vector of $K$ trainable coefficients $b^\gamma(\hat{u}) \in \mathbb{R}^K$, where $\gamma \in \{\alpha/2, 1-\alpha/2\}$. Similarly, the \textit{trunk} network consists of two components, one for each quantile. These two trunk components also have shared and independent trainable layers. They map the location $x \in K_2$ to a vector of $K$ basis functions $\tau^\gamma(x) \in \mathbb{R}^m,$ where $\gamma \in \{\alpha/2,1-\alpha/2\}$.

We can train the Quantile-DeepONet parameters $\theta = \{\theta_{\alpha/2},\theta_{1-\alpha/2}\}$ by minimizing the following loss function for $\gamma \in \{\alpha/2, 1-\alpha/2\}$:
$$
\mathcal{L}(\theta) = \frac{1}{N} \sum_{i=1}^N \rho_\gamma\left( G^{(i)},\hat{t}_{\theta_{\gamma}}\left(\hat{u}^{(i)}\right)\left(x^{(i)}\right)\right),
$$
on the dataset of $N$ triplets: $\mathcal{D} = \left\{\hat{u}^{(i)}, x^{(i)},G^{(i)} \right\}_{i=1}^N$. In the above expression, the loss function $\rho_\gamma$ refers to the pinball loss~\cite{koenker1978regression,steinwart2011estimating,romano2019conformalized}:
$$
\rho_\gamma(y,\hat{y}) = \begin{cases} \gamma(y-\hat{y}) \qquad \qquad \text{if } y-\hat{y}>0, \\  (1-\gamma)(\hat{y}-y) \quad~~ \text{otherwise}.\end{cases}
$$

Once trained, a Quantile-DeepONet can provide an estimate of the confidence interval for any new test sample:
$$
\hat{C}(\hat{u}_\text{test},x_\text{test}) = \left[\hat{t}_{\theta^*_{\alpha/2}}\left(\hat{u}_\text{test}\right)\left(x_\text{test}\right), \hat{t}_{\theta^*_{1-\alpha/2}}\left(\hat{u}_\text{test}\right)\left(x_\text{test}\right) \right],
$$
where $\theta^*_{\alpha/2}$ and $\theta^*_{1-\alpha/2}$ denote the optimizers of the corresponding losses.
However, the estimate provided above fails to meet the coverage probability requirement stated in~\eqref{eq:coverage-prob}. To satisfy this requirement, we need to incorporate conformal prediction. It is worth noting that the combination of quantile and conformal prediction is referred to as Conformal Quantile Regression (CQR) in the literature~\cite{romano2019conformalized}, and we will now describe its application to Quantile-DeepONets.

\textit{Conformal Quantile-DeepONet Regression.} The details of the split conformal Quantile-DeepONet regression procedure are provided in Algorithm~\ref{alg:CQR-DeepONet}. This procedure begins by defining a suitable score function.
\begin{algorithm}[t]
\DontPrintSemicolon
\caption{Split Conformal Quantile DeepONet Regression}
\label{alg:CQR-DeepONet}
\textbf{Input:} Calibration data: $\{\hat{u}^{(i)}, x^{(i)},G^{(i)}\}_{i=1}^{n}$, miscoverage level: $\alpha \in (0,1)$, and trained quantile-DeepONet outputs $\hat{t}_{\alpha/2}$ and $\hat{t}_{1-\alpha/2}$.\;
\textbf{Process:}\;
\Indp Use the trained quantile-DeepONet to compute the calibration scores: 
$$s_1 = s\left(\hat{u}^{(1)}, x^{(1)}, G^{(1)}\right), \dots, s_{n} = s\left(\hat{u}^{(n)}, x^{(n)}, G^{({n})}\right),$$ 
where 
$$s(u,x,G) = \max \left\{\hat{t}_{\theta^*_{\alpha/2}}(u)(x) - G, G- \hat{t}_{\theta^*_{1-\alpha/2}}(u)(x)\right\}.$$\;
Compute $\hat{q}$ as the $\frac{\lceil (n+1) (1-\alpha) \rceil}{n}$ quantile of the calibration scores $s_1, \dots, s_{n}$.\;
\Indm \textbf{Return:} Rigorous confidence intervals for any new test example $(\hat{u}_\text{test},x_{\text{test}})$:
$$\mathcal{C}(\hat{u}_\text{test},x_{\text{test}}) = \left[\hat{t}_{\theta^*_{\alpha/2}}(\hat{u}_\text{test})(x_\text{test})-\hat{q}, \hat{t}_{\theta^*_{1-\alpha/2}}(\hat{u}_\text{test})(x_\text{test}) + \hat{q}\right].$$
\end{algorithm}

\textit{Score function.} Given the trained Quantile-DeepONet, we compute the score for any DeepONet data triplet $(\hat{u}, x, G)$ as follows:
$$
s(u,x,G) = \max \left\{\hat{t}_{\theta^*_{\alpha/2}}(u)(x) - G, G- \hat{t}_{\theta^*_{1-\alpha/2}}(u)(x)\right\}.
$$
Note that if the operator target~$G$ satisfies~$G<\hat{t}_{\theta^*_{\alpha/2}}(u)(x)$ or $G> \hat{t}_{\theta^*_{1-\alpha/2}}(u)(x)$, then the score represents the amount of error incurred by Quantile-DeepONet.

\textit{Calibration quantile.} As before, using a given calibration i.i.d dataset of size $n$, denoted as $\{\hat{u}^{(i)}, x^{(i)}, G^{(i)}\}_{i=1}^N$, where $G^{(i)} = G(\hat{u}^{(i)})(x^{(i)})$, we compute the calibration scores $s_1 = s(\hat{u}^{(1)}, x^{(1)}, G^{(1)}), \dots, s_n = s(\hat{u}^{(n)}, x^{(n)}, G^{(n)})$. Then, we calculate $\hat{q}$ as the $\frac{\lceil (n+1) (1-\alpha) \rceil}{n}$  quantile of the calibration scores $s_1, \dots, s_n$. 

\textit{Confidence intervals with coverage guarantees.} We use the quantile $\hat{q}$ to construct rigorous confidence intervals for any new DeepONet test example $(\hat{u}_\text{test},x_{\text{test}})$:
\begin{align} \label{eq:confidence-intervals-CQR}
\mathcal{C}(\hat{u}_\text{test},x_{\text{test}}) = \left[\hat{t}_{\theta^*_{\alpha/2}}(\hat{u}_\text{test})(x_\text{test})-\hat{q}, \hat{t}_{\theta^*_{1-\alpha/2}}(\hat{u}_\text{test})(x_\text{test}) + \hat{q}\right].
\end{align}
These confidence intervals are adaptive, distribution-free, and satisfy the coverage probability property~\eqref{eq:coverage-prob}. The validity of this property was demonstrated in~\cite{romano2019conformalized} by extending theorem~\ref{theorem:conformal} to the case of CQR.

\section{Numerical Experiments} \label{sec:numerical-experiments}
In this section, we assess the ability of conformalized-DeepONets (conformal prediction combined with B-DeepONet, Prob-DeepONet, or Quantile-DeepONet) to generate confidence intervals with coverage guarantees. We illustrate this through three experiments: (i) the nonlinear pendulum (Section~\ref{subsec:pendulum}), (ii) the diffusion-reaction system~\ref{subsec:diffusion-reaction}, and (iii) the viscous Burgers' equation~\ref{subsec:viscous-burgers}.

\textbf{Datasets.} To train our DeepONet models, we used a dataset consisting of $N$ samples, each in the form of a triplet $\left(\hat{u}^{(i)},x^{(i)},G^{(i)}\right)$. The specific size of the training dataset, $N$, for each experiment is provided in Table~\ref{table:training-calibration-sizes}. For the calibration dataset, we used $n = \frac{N}{10}$ samples, with details outlined in Table~\ref{table:training-calibration-sizes}. Finally, for testing purposes, we evaluated all our DeepONet models using $100$ trajectories per experiment. Each trajectory included the target operator values $G(u)(X_\text{test})$ evaluated on a mesh $X_\text{test} \subset K_2$ composed of $100$ uniformly distributed location points within the output function domain $K_2$.
\textcolor{black}{ It should be noted that the testing mesh size is much more dense than the training mesh, i.e., our experiments also demonstrate the extrapolation of the proposed methods.}
\begin{table}[t]
\centering
\begin{tabular}{ccccc}
\hline
\textbf{Experiment} & \textbf{ \# $N$ training } & \textbf{\# $n$ calibration} \\
\hline
The nonlinear pendulum equation& 5000 & 500  \\
The diffusion-reaction system & 10000 & 1000  \\
The viscous Burgers' equation   & 30000 & 3000 \\
\hline
\end{tabular}
\caption{Summary of the number of training ($N$) and calibration ($n$) data samples for each experiment. Each sample is a triplet consisting of $\left(\hat{u}^{(i)},x^{(i)},G^{(i)}\right)$.}
\label{table:training-calibration-sizes}
\end{table}

\textbf{Neural Networks.} We used the feed-forward neural network (FNN) architecture to build the branch and trunk sub-networks. The corresponding width and depth for each experiment, including the shared and independent layers for Prob- and Quantile-DeepONet, are listed in Table~\ref{table:neural-nets}. We used the ReLU function as the activation function for each input and hidden layer.
\begin{table}[t]
\centering
\begin{tabular}{ccccc}
\toprule
\textbf{Experiment}  & \textbf{Trunk depth} & \textbf{Trunk width} & \textbf{Branch depth} & \textbf{Branch width} \\
\midrule
The nonlinear pendulum& 3(1) & 100 & 3(1) & 100 \\
The diffusion-reaction& 4(1) & 100 & 4(1) & 100 \\
The viscous Burgers'& 5(1) & 128 & 5(1) & 128 \\
\bottomrule
\end{tabular}
\caption{Summary of the width and depth of the trunk and branch feed-forward neural networks used in each experiment. For the depth of Prob- and Quantile-DeepONets, we use the notation \textbf{shared}(\textbf{independent}) layers.}
\label{table:neural-nets}
\end{table}

\textbf{Optimizer and Sampler.} For Prob- and Quantile-DeepONet, we used the Adam optimizer~\cite{kingma2014adam} with default hyperparameters and a learning rate~$\eta = 10^{-3}$. We adjusted the learning rate using the reduced on plateau scheduler. For B-DeepONet, we used the stochastic gradient replica-exchange sampling algorithm~\cite{lin2022multi,lin2023b,deng2020non} with standard hyperparameters.

\textbf{Baselines.} We compared the proposed Conformalized-DeepONets (which combine conformal prediction with B-DeepONet, Prob-DeepONet, or Quantile-DeepONet) against the following baselines: B-DeepONet~\cite{lin2023b}, Prob-DeepONet~\cite{moya2023deeponet}, and Quantile-DeepONet~(Section~\ref{subsec:CQR}). Specifically, for each conformalized and baseline model, we calculated the coverage over the test dataset. This coverage is determined by
$$
C_j = \frac{1}{n_\text{eval}} \sum_{i=1}^{n_\text{eval}} \mathbbm{1} \left \{ G^{(i,j)} \in \mathcal{C}_j \left(\hat{u}^{(j)}, x^{(i,j)} \right) \right\}, \text{ for } j=1,\dots,n_\text{traj}.
$$
In other words, we calculate the coverage of $n_\text{traj}$ test trajectories and \textcolor{black}{each $C_j$ is the ratio of location points whose reference solutions are in the established confidence interval.} These are evaluated at $n_{\text{eval}}=100$ uniformly spaced points within the output function domain. 

\subsection{Experiment: The Nonlinear Pendulum} \label{subsec:pendulum}
In this experiment, we consider the nonlinear pendulum subject to external forcing, represented by: 
\begin{align*}
    \frac{ds_1}{dt} &=s_2, \\
    \frac{ds_2}{dt} &=-k \sin s_1 + u(t),
\end{align*}
with the initial condition $(s_1(0),s_2(0))^\top = (0,0)^\top$ and time domain~$[0,1.0]$ (s). In the above, $(s_1(t),s_2(t))^\top$ is the state, $k=1.0$ is a constant determined by the acceleration due to gravity and the length of the pendulum, and $u(t)$ is the time-dependent external forcing with time domain~$[0,1]$ (s).

Our aim is to approximate the solution operator~$G:u(t) \mapsto s_1(t)$, which maps the external forcing input to the state $s_1$. We draw samples of the external forcing inputs for training, calibration, and testing from the mean-zero Gaussian Random Field~(GRF) $u \sim \mathcal{G}(0,k_\ell(t_1, t_2))$. This GRF has a radial-basis function covariance kernel $k_\ell(t_1,t_2)$ with a length-scale of $\ell=0.1$. Finally, we discretize each sampled input~$u$ using $m=100$ sensors.

\begin{figure}[t!] 
    \begin{subfigure}[!bl]{0.50\textwidth}
        \begin{tikzpicture}
             \begin{axis}[ xlabel={$t$}, ylabel={$G(u)(t) \equiv s_1(t)$},xmin=0, xmax=1,ymin=-0.07, ymax=0.01, legend pos=south west, line width=1pt,enlargelimits=false]     
             
                    \addplot[solid, color=blue]  
                            table{figsData/Pendulum/pendulum_true_traj_prob36.txt};
                                \addlegendentry{Reference Solution}
                     \addplot[name path=prob_lower,  fill=none, draw=none,forget plot]                 
                             table{figsData/Pendulum/pendulum_lower_traj_prob36.txt};
                      \addplot[name path=prob_upper, fill=none, draw=none,forget plot] 
                             table{figsData/Pendulum/pendulum_upper_traj_prob36.txt};
                             
                     \addplot[name path=cop_lower, fill=none, draw=none,forget plot]                 
                             table{figsData/Pendulum/pendulum_lower_traj_cop36.txt};
                      \addplot[name path=cop_upper, fill=none, draw=none,forget plot] 
                             table{figsData/Pendulum/pendulum_upper_traj_cop36.txt};

                     \definecolor{mycolor1}{RGB}{75,200,75}
                     \addplot[fill=mycolor1, fill opacity=0.5] fill between[of=cop_upper and cop_lower, ];  
                     \addlegendentry{Conformalized-DeepONet}
                     
                     \definecolor{mycolor2}{RGB}{0,255,255}
                     \addplot[fill=red, fill opacity=0.5] fill between[of=prob_upper and prob_lower, ];
                     \addlegendentry{Prob-DeepONet}                               
         \end{axis}
        \end{tikzpicture}
    \caption{  Conformal Prediction and Prob-DeepONet}
    \label{subfig:Probabilistic_pendulum}
    \end{subfigure}
    \hspace{10pt}
    \begin{subfigure}[!br]{0.5\textwidth}
        \begin{tikzpicture}
             \begin{axis}[ xlabel={$t$}, ylabel={$G(u)(t) \equiv s_1(t)$},xmin=0, xmax=1,ymin=-0.07, ymax=0.01, legend pos=south west, line width=1pt,enlargelimits=false]     
                    \addplot[solid, color=blue]  
                            table{figsData/Pendulum/pendulum_true_traj_quantile36.txt};
                                \addlegendentry{Reference Solution}
                     \addplot[name path=quantile_lower,  fill=none, draw=none,forget plot]     
                             table{figsData/Pendulum/pendulum_lower_traj_quantile36.txt};
                      \addplot[name path=quantile_upper, fill=none, draw=none,forget plot] 
                             table{figsData/Pendulum/pendulum_upper_traj_quantile36.txt};
                             
                     \addplot[name path=coq_lower, fill=none, draw=none,forget plot]           
                             table{figsData/Pendulum/pendulum_lower_traj_coq36.txt};
                      \addplot[name path=coq_upper, fill=none, draw=none,forget plot] 
                             table{figsData/Pendulum/pendulum_upper_traj_coq36.txt};
                     \definecolor{mycolor1}{RGB}{75,200,75}
                     \addplot[fill=mycolor1, fill opacity=0.5] fill between[of=coq_upper and coq_lower, ];  
                     \addlegendentry{Conformalized-DeepONet}
                     
                     \definecolor{mycolor2}{RGB}{0,255,255}
                     \addplot[fill=red, fill opacity=0.5] fill between[of=quantile_upper and quantile_lower, ];
                     \addlegendentry{Quantile-DeepONet}                               
         \end{axis}
        \end{tikzpicture}
    \caption{Conformal Prediction and Quantile-DeepONet}
    \label{subfig: quantalize_pendulum}
    \end{subfigure}
\caption{Confidence intervals for a random test trajectory of the nonlinear pendulum experiment given a miscoverage rate $\alpha=0.05$. \textbf{(a)} Confidence intervals for the Conformalized Prob-DeepONet and the baseline Prob-DeepONet. \textbf{(b)} Confidence intervals for the Conformalized Quantile-DeepONet and the baseline Quantile-DeepONet. Please refer to Table \ref{table:pendulum-coverage} for the summary of the average coverage rate for all testing trajectories with different approaches. }
\label{fig:pendulum-CIs}
\end{figure}
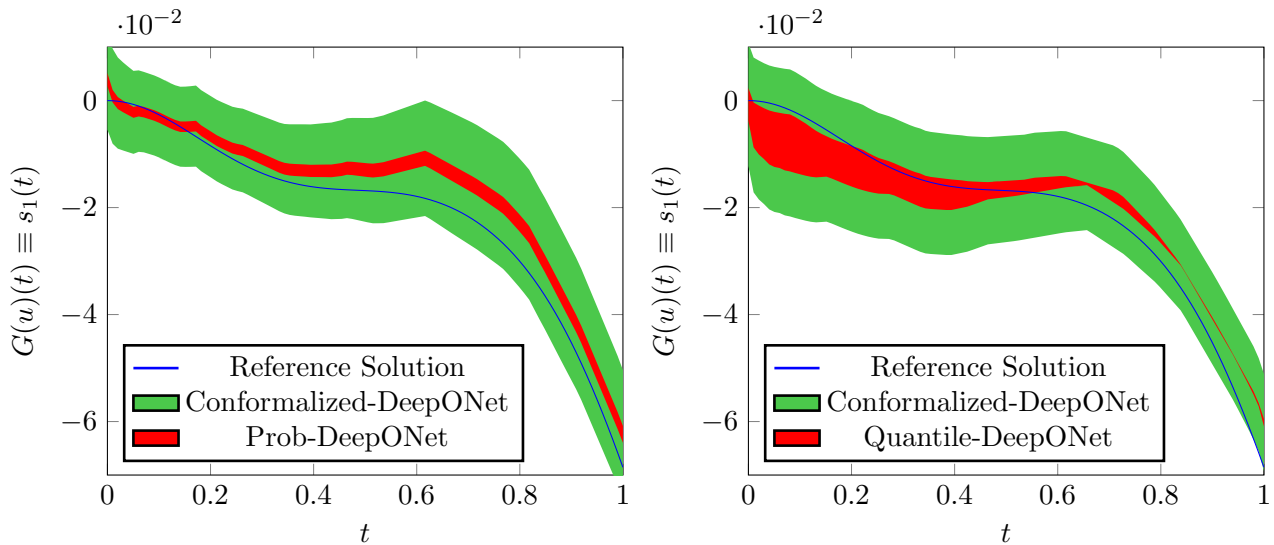

\textbf{Checking Coverage.} We evaluated the ability of all conformalized-DeepONet models and baseline models to generate confidence intervals (with misscoverage rate $\alpha=0.05$) with coverage guarantees for trajectories in the test dataset. Figure~\ref{fig:pendulum-CIs} illustrates the confidence intervals for a test trajectory randomly selected from the test dataset. Furthermore, Table~\ref{table:pendulum-coverage} presents the average coverage achieved by both conformalized-DeepONets and baseline models for all trajectories in the test dataset. 

\begin{table}[t]
\centering
\begin{tabular}{c   c   } 
\hline
\textbf{DeepONet model} &\textbf{Average coverage \% ($\alpha=0.05$)} \\
\hline
Conformalized Prob-DeepONet & \textbf{94.69}\%  \\
Conformalized Quantile-DeepONet & \textbf{95.86}\%   \\
\hline
Prob-DeepONet & 3.80\% \\
Quantile-DeepONet & 55.31\%   \\
\hline
\end{tabular}
\caption{The average coverage percentage for the predicted confidence intervals (with a miscoverage rate of $\alpha=0.05$) calculated based on the 100 test trajectories in the nonlinear pendulum experiment.}
\label{table:pendulum-coverage}
\end{table}

Figure~\ref{fig:pendulum-CIs} and Table~\ref{table:pendulum-coverage} offer significant insights. Firstly, the conformalized DeepONets (both Probabilistic and Quantile) deliver the desired coverage on average, with a miscoverage rate of $\alpha=0.05$. Moreover, the Quantile-DeepONet significantly outperforms the Probabilistic DeepONet. Lastly, the Probabilistic DeepONet displays the poorest average coverage. These findings corroborate the conclusions from our previous paper, which highlighted the necessity of extensive hyperparameter optimization to improve the quality of confidence intervals for Prob-DeepONet.

\begin{figure}[t]
  \centering
  \begin{tikzpicture}
             \begin{axis}[width=\linewidth*0.7, xlabel={Predicted confidence intervals' coverage (\%)}, ylabel={Conuts}, ymax=400,
             ]     
             \addplot [black, dashed] coordinates {(0.95,0) (0.95,350)};
            \addlegendentry{$1-\alpha$} 
            \addplot [
                hist={
                    bins=10, 
                },
             fill=cyan,
             mark=none,
            ] table [y index=0] {figsData/statsData/coverages.txt}; %
         \addplot [black, dashed] coordinates {(0.95,0) (0.95,450)};

            \end{axis}
  \end{tikzpicture}
  \caption{The distribution of predicted confidence intervals' coverages (in percentages), obtained by using the conformalized Quantile-DeepONet on the 100 test trajectories of the nonlinear pendulum experiment, with a specified miscoverage rate of $\alpha=0.05$.}
  \label{fig:distr-coverages}
\end{figure}
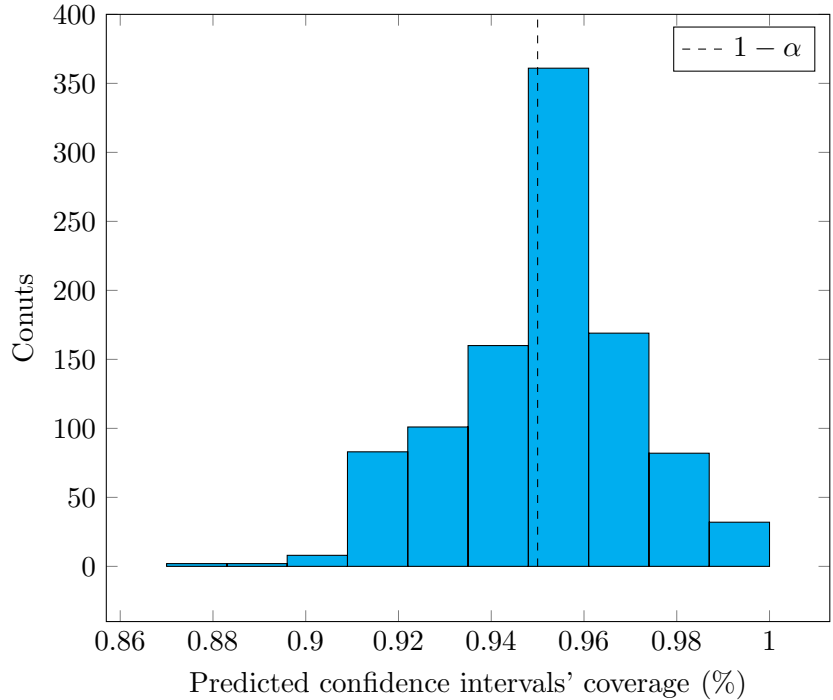

Finally, Figure~\ref{fig:distr-coverages} shows the distribution of coverages (with a miscoverage rate of $\alpha=0.05$) for all confidence intervals obtained by applying the conformalized Quantile-DeepONet to all $100$ test trajectories of the nonlinear pendulum example. As anticipated, the coverages concentrate around $1-\alpha$, that is, with 95\% coverage. Next, we present two ablation studies showing the adaptivity and effect of the calibration dataset size.

\begin{figure}[t]
  \centering
  \begin{tikzpicture}
             \begin{axis}[width=\linewidth*0.7, xlabel={Confidence intervals' lengths ($\alpha=0.05$)}, ylabel={Counts},     
             xtick={0, 0.02, 0.04, 0.06, 0.08, 0.1, 0.12, 0.14},
             xticklabels={0, $0.02$, $0.04$, $0.06$, $0.08$, $0.1$, $0.12$, $0.14$},
             ] 
            \addplot +[
                hist={
                    bins=20, 
                },
             fill=cyan,
             mark=none,
             draw=black,
            ] table [y index=0] {figsData/statsData/set_sizes.txt}; 
            \end{axis}
  \end{tikzpicture}
\caption{The distribution of the lengths of the confidence intervals (given a miscoverage rate $\alpha=0.05$) predicted using the proposed conformalized Prob-DeepONet for the $100$ test trajectories of the nonlinear pendulum experiment.}
  \label{fig:intervals-lengths}
\end{figure}
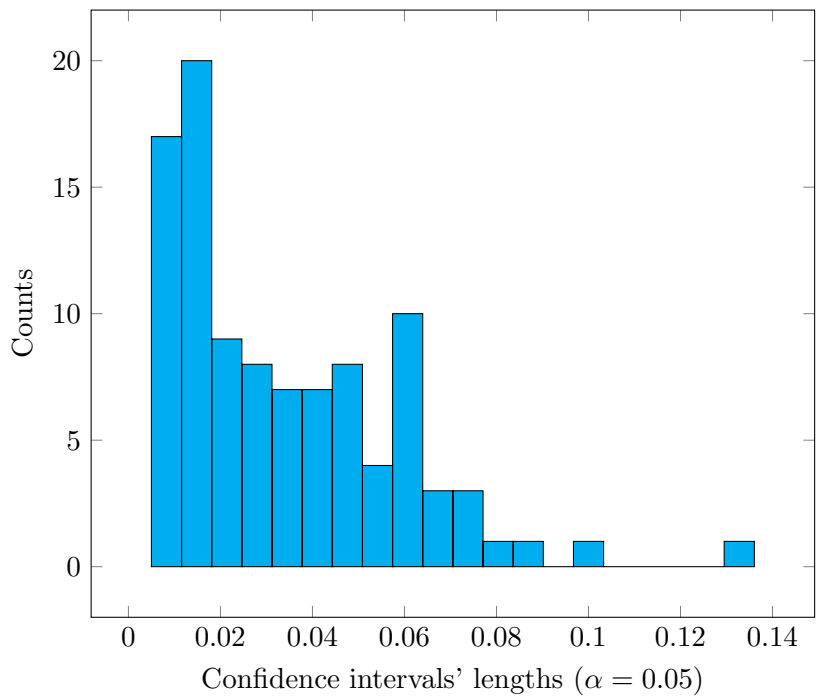

\textbf{Ablation Study I: Adaptivity of Confidence Intervals.} We tested the proposed conformalized Prob-DeepONet for its ability to provide adaptive confidence intervals -- that is, confidence intervals of varying lengths for each $x \in K_2$, while still ensuring coverage guarantees. 
The adaptive coverage will reduce the over-confidence issue of many algorithms.
To accomplish this, we applied the conformalized Prob-DeepONet to all $100$ test trajectories of the pendulum experiment, with the goal of analyzing the distribution of the confidence intervals' lengths. Figure~\ref{fig:intervals-lengths} illustrates the distribution of the lengths of confidence intervals. This distribution demonstrates that the confidence intervals are adaptive, meaning their length varies for each $x \in K_2$. If this wasn't the situation, the distribution would be focused around a specific length. Clearly, our experiment indicates that this is not the scenario.

\textbf{Ablation Study II: Testing the Effect of the Calibration Dataset Size~$n$.} We tested the effect of varying the calibration dataset size~$n$ on the distribution of coverages for the nonlinear pendulum example and a miscoverage rate of $\alpha=0.05$. 

To this end, we varied $n$ over the set $\{500, 1000, 5000, 10000\}$. For each $n$, we computed the distribution of coverages. This was done for $R$ rounds ($R=200$), where we generated a calibration dataset of size $n$ and a validation dataset of size $n_\text{val}$. Each dataset contains DeepONet triplets of the form $(\hat{u}, x, G)$. We then calculated the coverages as follows. 
$$
C_k = \frac{1}{n_\text{val}} \sum_{i=1}^{n_\text{val}} \mathbbm{1} \left\{ G^{(i,k)} \in \mathcal{C}_k\left(\hat{u}^{(i,k)}, x^{(i,k)}\right)\right\}, \quad \text{for } k=1,\dots,R.
$$
Finally, we obtained the distribution of the obtained coverages $\{C_k\}_{k=1}^R,$ which we plot for each $n$.

Note that to prevent the generation of $R$ distinct datasets of size $n+n_\text{val}$, we used the approach proposed in~\cite{angelopoulos2021gentle}. This method creates just one dataset of size $n+n_\text{val}$ for all $R$ rounds. Then, in each round, we shuffled this dataset and used the initial $n$ samples for calibration and the remaining $n_\text{val}$ samples for validation. 

\begin{figure}[!h]
  \centering
  \begin{tikzpicture}
             \begin{axis}[width=\linewidth*0.7, xlabel={Average coverage $C_k$}, ylabel={Counts}, ymin=0, ymax=70,legend pos=north west]     
                      \addplot [dashed,thick, color=black] coordinates {(0.95,0) (0.95,70)};
                                  \addlegendentry{$1-\alpha$} 
                    \addplot[solid, color=blue,line width=1pt]  
                            table{figsData/statsData/500data.txt};
                                \addlegendentry{n=500}
                    \addplot[solid, color=orange,line width=1pt]  
                            table{figsData/statsData/1000data.txt};
                                \addlegendentry{n=1000}
                    \addplot[solid, color=red,line width=1pt]  
                            table{figsData/statsData/5000data.txt};
                                \addlegendentry{n=5000}
                    \definecolor{mycolor1}{RGB}{75,200,75}
                    \addplot[solid, color=mycolor1,line width=1pt]  
                            table{figsData/statsData/10000data.txt};
                                \addlegendentry{n=10000}
            \end{axis}
  \end{tikzpicture}
  \caption{The empirical distribution of coverages~$C_k$ for different values of the calibration dataset size~$n \in \{500, 1000, 5000, 10000\}$. We observe that for $n>500$, the distribution, as expected, centers around $95$\% ($\alpha=0.05$). Notably, for $n=1000$ and this nonlinear pendulum dataset, we achieve a satisfactory distribution, and the gain from increasing $n$ is minimal. }
  \label{fig:distr-coverages_with_different_n}
\end{figure}
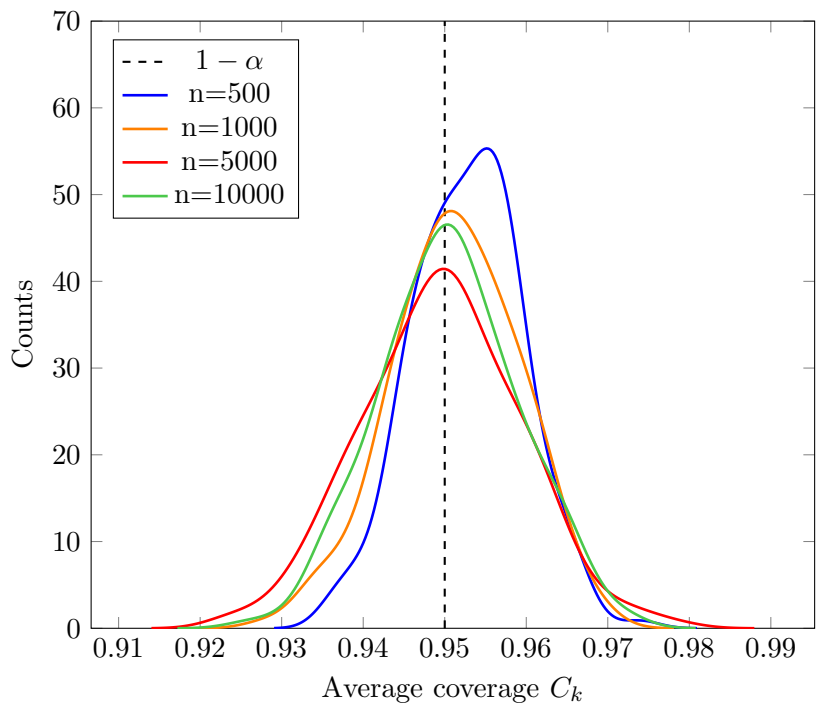

Figure~\ref{fig:distr-coverages_with_different_n} illustrates the obtained distribution of coverages for $n \in \{500, 1000, 5000, 10000\}$. Observe that for $n>500$, the distribution, as expected, centers around 95\% ($\alpha=0.05$). Notably, for $n=1000$, we achieve a satisfactory distribution, and the gain from increasing the calibration dataset size is minimal.

Finally, for $n=500$, the distribution peaks at a value larger than 95\%. This could be due to the small number of calibration samples, which may not allow for a proper characterization of the coverage. However, even in this extreme case, conformalized-DeepONet still produces satisfactory confidence intervals around $1-\alpha$.

\subsection{Experiment: The Diffusion-Reaction System} \label{subsec:diffusion-reaction}
In the second experiment, we examine the following diffusion-reaction system with a source term, $u(x)$, as described below:
\begin{align*}
& \frac{\partial s}{\partial t} = D \frac{\partial^2 s}{\partial x^2} + k s^2 + u(x), \qquad x \in [0,1],~t\in[0,1],
\end{align*}
with zero boundary/initial conditions. In the above, $D=0.01$ is the diffusion coefficient, and $k=0.01$ is the reaction rate.

Our goal is to approximate the solution operator $G:u(x) \mapsto s(x,1.0)$, that is, the mapping from the source term to the solution at the terminal time. As in the nonlinear pendulum experiment, we sample the input function (the source term $u(x)$) from the non-zero Gaussian Random Field $u \sim \mathcal{G}(0, k_\ell(x_1, x_2))$ with a radial basis function kernel and a length-scale of $\ell=0.1$. Finally, we discretize the sampled source terms using $m=100$ sensors.

\begin{figure}[t] 
    \begin{subfigure}[!bl]{0.50\textwidth}
        \begin{tikzpicture}
             \begin{axis}[ xlabel={$x$}, ylabel={$G(u)(x) \equiv s(x,1.0)$},xmin=0, xmax=1,ymin=-0.7, ymax=3.1, legend pos=north west, line width=1pt,enlargelimits=false]     
             
                    \addplot[solid, color=blue]  
                            table{figsData/Diffusion/diffusion_true_traj_prob43.txt};
                                \addlegendentry{Reference Solution}
                     \addplot[name path=prob_lower,  fill=none, draw=none,forget plot]                 
                             table{figsData/Diffusion/diffusion_lower_traj_prob43.txt};
                      \addplot[name path=prob_upper, fill=none, draw=none,forget plot] 
                             table{figsData/Diffusion/diffusion_upper_traj_prob43.txt};
                             
                     \addplot[name path=cop_lower, fill=none, draw=none,forget plot]                 
                             table{figsData/Diffusion/diffusion_lower_traj_cop43.txt};
                      \addplot[name path=cop_upper, fill=none, draw=none,forget plot] 
                             table{figsData/Diffusion/diffusion_upper_traj_cop43.txt};

                     \definecolor{mycolor1}{RGB}{75,200,75}
                     \addplot[fill=mycolor1, fill opacity=0.5] fill between[of=cop_upper and cop_lower, ];  
                     \addlegendentry{Conformalized-DeepONet}
                     
                     \definecolor{mycolor2}{RGB}{0,255,255}
                     \addplot[fill=red, fill opacity=0.5] fill between[of=prob_upper and prob_lower, ];
                     \addlegendentry{Prob-DeepONet}                               
         \end{axis}
        \end{tikzpicture}
    \caption{Conformal Prediction and Prob-DeepONet}
    \label{subfig:Probabilistic_Diffusion}
    \end{subfigure}
    \hspace{10pt}
    \begin{subfigure}[!br]{0.5\textwidth}
        \begin{tikzpicture}
             \begin{axis}[ xlabel={$x$}, ylabel={$G(u)(x) \equiv s(x,1.0)$},,xmin=0, xmax=1,ymin=-0.7, ymax=3.1, legend pos=north west, line width=1pt,enlargelimits=false]     
                    \addplot[solid, color=blue]  
                            table{figsData/Diffusion/diffusion_true_traj_quantile43.txt};
                                \addlegendentry{Reference Solution}
                     \addplot[name path=quantile_lower,  fill=none, draw=none,forget plot]     
                             table{figsData/Diffusion/diffusion_lower_traj_quantile43.txt};
                      \addplot[name path=quantile_upper, fill=none, draw=none,forget plot] 
                             table{figsData/Diffusion/diffusion_upper_traj_quantile43.txt};
                             
                     \addplot[name path=coq_lower, fill=none, draw=none,forget plot]           
                             table{figsData/Diffusion/diffusion_lower_traj_coq43.txt};
                      \addplot[name path=coq_upper, fill=none, draw=none,forget plot] 
                             table{figsData/Diffusion/diffusion_upper_traj_coq43.txt};
                     \definecolor{mycolor1}{RGB}{75,200,75}
                     \addplot[fill=mycolor1, fill opacity=0.5] fill between[of=coq_upper and coq_lower, ];  
                     \addlegendentry{Conformalized-DeepONet}
                     
                     \definecolor{mycolor2}{RGB}{0,255,255}
                     \addplot[fill=red, fill opacity=0.5] fill between[of=quantile_upper and quantile_lower, ];
                     \addlegendentry{Quantile-DeepONet}                               
         \end{axis}
        \end{tikzpicture}
    \caption{Conformal Prediction and Quantile-DeepONet}
    \label{subfig: quantalize_Diffusion}
    \end{subfigure}
\caption{Confidence intervals for a random test trajectory of the diffusion-reaction experiment given a miscoverage rate $\alpha=0.05$. \textbf{(a)} Confidence intervals for the Conformalized Prob-DeepONet and the baseline Prob-DeepONet. \textbf{(b)} Confidence intervals for the Conformalized Quantile-DeepONet and the baseline Quantile-DeepONet. \textcolor{black}{The average coverage rate for all samples for all approaches are presented in Table \ref{table:diffusion-reaction-coverage}.} }
\label{fig:diffusion-reaction-CI}
\end{figure}
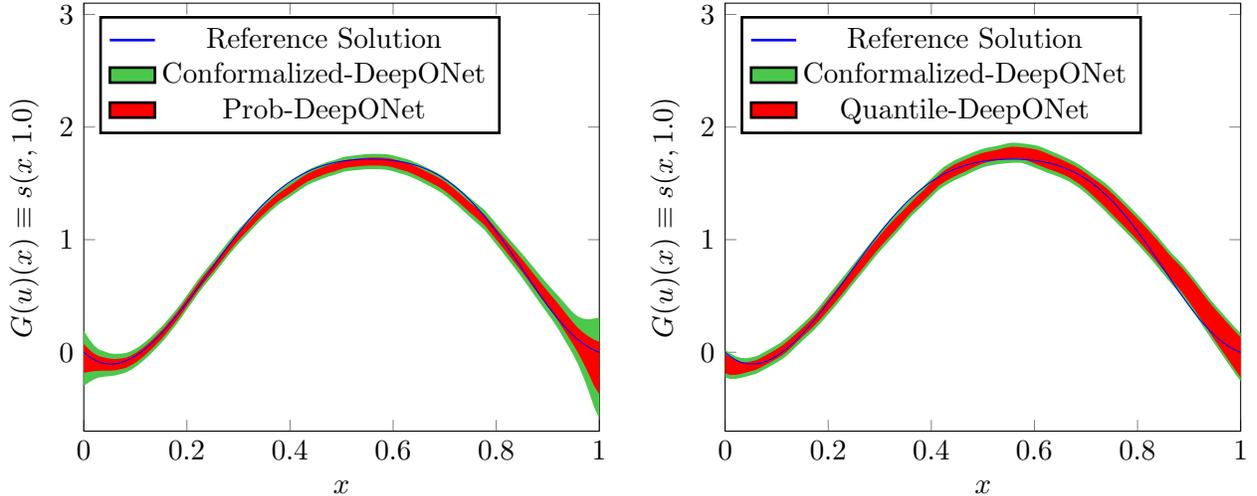

\textbf{Checking Coverage.} We assessed the ability of all conformalized and baseline DeepONet models to generate appropriate confidence intervals for a miscoverage rate of $\alpha=0.05$. Figure~\ref{fig:diffusion-reaction-CI} displays the confidence intervals predicted by the proposed conformalized-DeepONet and baseline models for a randomly selected trajectory from the test dataset. Furthermore, Table~\ref{table:diffusion-reaction-coverage} presents the average coverage of the confidence intervals predicted by both the conformalized and baseline DeepONet models for all $n_\text{traj}=100$ test trajectories of the diffusion-reaction system.  

Figure~\ref{fig:diffusion-reaction-CI} shows that the confidence intervals predicted by the conformalized-DeepONet appear to be adequate. Table~\ref{table:diffusion-reaction-coverage} confirms this, indicating that the average coverage for the conformalized-DeepONets is as expected ($\approx 95$\%), considering a miscoverage rate of $\alpha=0.05$. The baseline Quantile-DeepONet also has a good coverage level compared to the Prob-DeepONet. This superior coverage might enable the conformalized Quantile-DeepONet to perform slightly better than the conformalized Prob-DeepONet, which must compensate for its baseline's poor coverage.

\begin{table}[t]
\centering
\begin{tabular}{c   c   } 
\hline
\textbf{DeepONet model} &\textbf{Average coverage \% ($\alpha=0.05$)} \\
\hline
Conformalized Prob-DeepONet & \textbf{96.26}\%  \\
Conformalized Quantile-DeepONet & \textbf{95.64}\%   \\
\hline
Prob-DeepONet & 8.99\% \\
Quantile-DeepONet & 83.09\%   \\
\hline
\end{tabular}
\caption{The average coverage percentage for the predicted confidence intervals (with a miscoverage rate of $\alpha=0.05$) calculated based on the 100 test trajectories in the diffusion-reaction experiment.}
\label{table:diffusion-reaction-coverage}
\end{table}

\subsection{Experiment: The Viscous Burgers' Equation} \label{subsec:viscous-burgers}
In this section, we explore a common example found in most operator learning literature: the viscous Burgers' equation:
\begin{align*}
    &\frac{\partial u_s}{\partial t} + \frac{1}{2}\frac{\partial (u^2_s)}{\partial x} = \alpha \frac{\partial^2 u_s}{\partial x^2},\hspace{0.5em} x\in[0, 2\pi], \hspace{0.2em} t\in[0, 0.3]\\
    &u_s(x, 0) = u^0_s(x),\\
    &u_s(0, t) = u_s(2\pi, t),
\end{align*}
where $u^0_s(x)$ is the initial condition that depends on the parameter $s$. In the above, the viscosity is set to $\alpha = 0.05$. 

Our objective is to estimate the solution operator $G:u_{s}^0(x) \mapsto u_s(x,0.3)$, which is the mapping from the initial condition to the solution at the terminal time. We generated initial conditions using a sum of two Gaussian distributions with a uniformly randomized weight from $[0, 5]$. The means and standard deviations of the two distributions were sampled uniformly from $[0, 2\pi]$ and $[0.1, 1]$, respectively. 

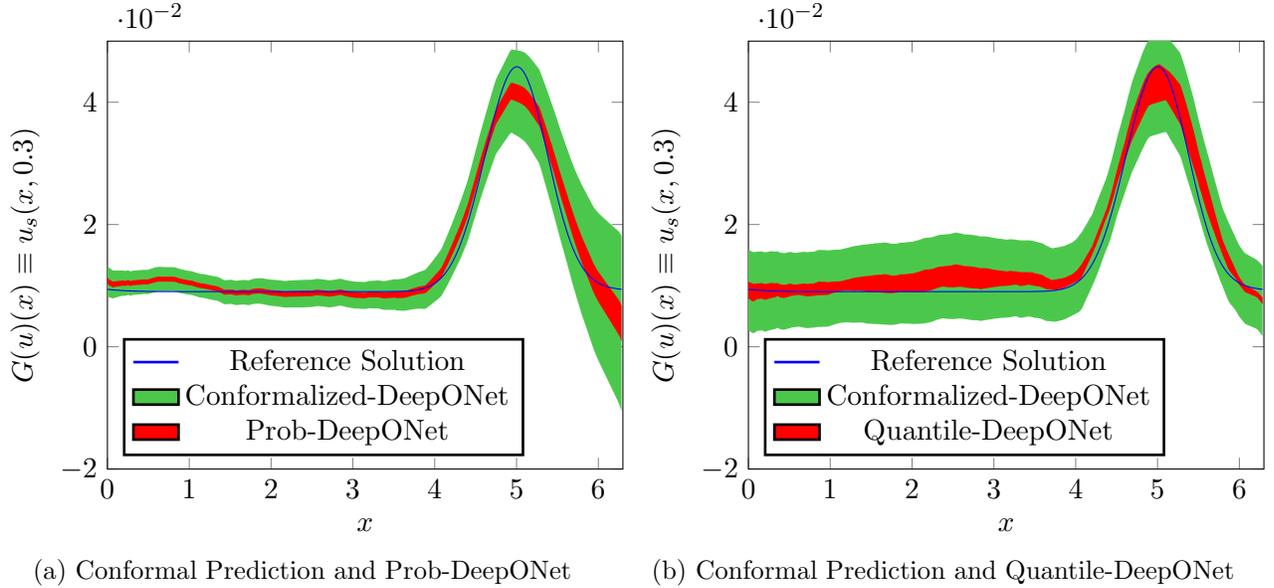
\begin{figure}[t] 
    \begin{subfigure}[!bl]{0.50\textwidth}
        \begin{tikzpicture}
             \begin{axis}[ xlabel={$x$}, ylabel={$G(u)(x) \equiv u_s(x,0.3)$}, ,xmin=0, xmax=6.3,ymin=-0.02, ymax=0.05,legend pos=south west, line width=1pt,enlargelimits=false]     
             
                    \addplot[solid, color=blue]  
                            table{figsData/Burgers/Burgers_true_traj_prob56.txt};
                                \addlegendentry{Reference Solution}
                     \addplot[name path=prob_lower,  fill=none, draw=none,forget plot]                 
                             table{figsData/Burgers/Burgers_lower_traj_prob56.txt};
                      \addplot[name path=prob_upper, fill=none, draw=none,forget plot] 
                             table{figsData/Burgers/Burgers_upper_traj_prob56.txt};
                             
                     \addplot[name path=cop_lower, fill=none, draw=none,forget plot]                 
                             table{figsData/Burgers/Burgers_lower_traj_cop56.txt};
                      \addplot[name path=cop_upper, fill=none, draw=none,forget plot] 
                             table{figsData/Burgers/Burgers_upper_traj_cop56.txt};

                     \definecolor{mycolor1}{RGB}{75,200,75}
                     \addplot[fill=mycolor1, fill opacity=0.5] fill between[of=cop_upper and cop_lower, ];  
                     \addlegendentry{Conformalized-DeepONet}
                     
                     \definecolor{mycolor2}{RGB}{0,255,255}
                     \addplot[fill=red, fill opacity=0.5] fill between[of=prob_upper and prob_lower, ];
                     \addlegendentry{Prob-DeepONet}                               
         \end{axis}
        \end{tikzpicture}
    \caption{Conformal Prediction and Prob-DeepONet}
    \label{subfig:Probabilistic_Burger}
    \end{subfigure}
    \hspace{10pt}
    \begin{subfigure}[!br]{0.5\textwidth}
        \begin{tikzpicture}
             \begin{axis}[ xlabel={$x$}, ylabel={$G(u)(x) \equiv u_s(x,0.3)$}, xmin=0, xmax=6.3,ymin=-0.02, ymax=0.05,legend pos=south west, line width=1pt,enlargelimits=false]     
             
                    \addplot[solid, color=blue]  
                            table{figsData/Burgers/Burgers_true_traj_quantile56.txt};
                                \addlegendentry{Reference Solution}
                     \addplot[name path=quantile_lower,  fill=none, draw=none,forget plot]                 table{figsData/Burgers/Burgers_lower_traj_quantile56.txt};
                      \addplot[name path=quantile_upper, fill=none, draw=none,forget plot] 
                             table{figsData/Burgers/Burgers_upper_traj_quantile56.txt};
                             
                     \addplot[name path=coq_lower, fill=none, draw=none,forget plot]                 
                             table{figsData/Burgers/Burgers_lower_traj_coq56.txt};
                      \addplot[name path=coq_upper, fill=none, draw=none,forget plot] 
                             table{figsData/Burgers/Burgers_upper_traj_coq56.txt};

                     \definecolor{mycolor1}{RGB}{75,200,75}
                     \addplot[fill=mycolor1, fill opacity=0.5] fill between[of=coq_upper and coq_lower, ];  
                     \addlegendentry{Conformalized-DeepONet}
                     
                     \definecolor{mycolor2}{RGB}{0,255,255}
                     \addplot[fill=red, fill opacity=0.5] fill between[of=quantile_upper and quantile_lower, ];
                     \addlegendentry{Quantile-DeepONet}                               
         \end{axis}
        \end{tikzpicture}
    \caption{Conformal Prediction and Quantile-DeepONet}
    \label{subfig: quantalize_Burger}
    \end{subfigure}
\caption{Confidence intervals for a random test trajectory of the viscous Burgers' experiment given a miscoverage rate $\alpha=0.05$. \textbf{(a)} Confidence intervals for the Conformalized Prob-DeepONet and the baseline Prob-DeepONet. \textbf{(b)} Confidence intervals for the Conformalized Quantile-DeepONet and the baseline Quantile-DeepONet. The average coverage for all testing trajectories and approaches are summarized and presented in Table \ref{table:viscous-burgers-coverage}. }
\label{fig:viscous-Burgers}
\end{figure}

\textbf{Checking Coverage.} We evaluated the proposed conformalized-DeepONet models' capacity to produce rigorous confidence intervals. These intervals have a specified miscoverage rate of $\alpha=0.05$ for test trajectories generated from the viscous Burgers' model. We also compared these proposed models with the baseline models. Figure~\ref{fig:viscous-Burgers} presents a comparison between the conformalized and baseline models. This comparison uses the predicted confidence interval for a randomly chosen test trajectory from the viscous Burgers' example. The figure indicates that both the conformalized Prob-DeepONet and Quantile-DeepONet effectively capture the reference solution within the predicted confidence interval. In constrast, the baseline models do not capture several portions of the reference solution. 

Moreover, Table~\ref{table:viscous-burgers-coverage} shows the average coverage of confidence intervals (with a targeted coverage rate of $\alpha=0.05$) generated from both conformalized and baseline models for all $n_\text{traj}=100$ test trajectories produced using the viscous Burgers' model. The results once again demonstrate that both the conformalized Prob-DeepONet and the conformalized Quantile-DeepONet provide (on average) the desired coverage guarantee. In contrast, both baseline models generally offer poor coverage guarantees. It's noteworthy that the conformalized Prob-DeepONet still produces satisfactory coverage results, even though these are based on the inferior results from the baseline Prob-DeepONet model.

\begin{table}[t]
\centering
\begin{tabular}{c   c   } 
\hline
\textbf{DeepONet model} &\textbf{Average coverage \% ($\alpha=0.05$)} \\
\hline
Conformalized Prob-DeepONet & \textbf{94.88}\%  \\
Conformalized Quantile-DeepONet & \textbf{95.96}\%   \\
\hline
Prob-DeepONet & 7.84\% \\
Quantile-DeepONet & 72.91\%   \\
\hline
\end{tabular}
\caption{The average coverage percentage for the predicted confidence intervals (with a miscoverage rate of $\alpha=0.05$) calculated based on the 100 test trajectories in the viscous Burgers' experiment.}
\label{table:viscous-burgers-coverage}
\end{table}

\section{Discussion} \label{sec:discussion}

\textbf{On Our Results.} Our results indicate that split conformal prediction~\cite{vovk2005algorithmic,angelopoulos2021gentle,romano2019conformalized} is an easy-to-use methodology that offers confidence intervals with coverage guarantees. We consider it easy-to-use as it does not require any distributional assumptions and only needs a finite number of DeepONet samples. This allowed us to build an adaptive, reliable, and efficient Uncertainty Quantification (UQ) and regression framework for Deep Operator Networks.

Our results also indicate that split conformal prediction is model agnostic. Note that we  applied it on top of two UQ DeepONet frameworks (B-DeepONet and Prob-DeepONet), proposed in our previous works, and a new UQ framework (Quantile-DeepONet) proposed here. Hence, we expect split conformal prediction to be compatible with many existing DeepONet extensions, such as Multi-fidelity DeepONet~\cite{lu2022multifidelity,howard2022multifidelity,mollaali2023physics}, BelNet~\cite{zhang2022belnet}, Fed-DeepONet~\cite{moya2022fed}, or D2NO~\cite{zhang2023d2no}. To demonstrate this, we will present a simple experiment that highlights the use of split conformal prediction with Multi-fidelity DeepONets. 

\textbf{Conformal Prediction and Multi-Fidelity DeepONet.}
In this experiment, we used a multi-fidelity conformalized Prob-DeepONet model to approximate the high-fidelity solution operator for the given 1D jump function:
\begin{equation}
\begin{aligned}
y_L(u)(x) &= 
\begin{cases} 
0.5(6x - 2)^2 \sin(u) + 10(x - 0.5) - 5 &  x \leq 0.5 \\
0.5(6x - 2)^2 \sin(u) + 10(x - 0.5) - 2 &  x > 0.5
\end{cases} \\
y_H(u)(x) &= 2y_L(u)(x) - 20x + 20 \\
u(x) &= ax - 4
\end{aligned}
\end{equation}
where $x \in [0, 1]$ and a $\in [10, 14]$. In the above, $y_L$ is the low-fidelity solution, $y_H$ the high-fidelity solution, and $u$ the input.

We built our Multi-fidelity Prob-DeepONet baseline model as follows. Initially, we used a trained classical DeepONet to approximate the low-fidelity solution operator $G:u(x) \mapsto \hat{y}{LF}(u)(x)$. Next, we trained a Prob-DeepONet to approximate the difference between the high-fidelity and low-fidelity solution operators $(y_{HF} - y_{LF})~|~X=(u,x) \sim \mathcal{N}(\mu(u)(x),\sigma(u)(x))$. Finally, we recovered the high-fidelity solution operator using the probabilistic model~$\mathcal{N}(\mu(u)(x) + \hat{y}(u)(x), \sigma(u)(x))$.

We trained the baseline model using a dataset of $N_{LF}=3800$ low-fidelity DeepONet triplets and $N_{HF}=760$ high-fidelity DeepONet triplets. For calibration, we used the half the of number of high-fidelity data samples ($n=\frac{N_{HF}}{2} = 380$). 

\begin{figure}[h]
  \centering
  \begin{tikzpicture}
             \begin{axis}[width=\linewidth*0.7, xlabel={$x$}, ylabel={$y_H(u)(x)$},ymin=-4.5, ymax=14.0,legend pos=north west]     
             
                    \addplot[dashed, color=blue,mark= *, mark size=1.2pt]  
                            table{figsData/MF/MF_true_traj_prob97.txt};
                                \addlegendentry{High-fidelity Reference Solution}
                     \addplot[name path=prob_lower,  fill=none, draw=none,forget plot]                 
                             table{figsData/MF/MF_lower_traj_prob97.txt};
                      \addplot[name path=prob_upper, fill=none, draw=none,forget plot] 
                             table{figsData/MF/MF_upper_traj_prob97.txt};
                             
                     \addplot[name path=cop_lower, fill=none, draw=none,forget plot]                 
                             table{figsData/MF/MF_lower_traj_cop97.txt};
                      \addplot[name path=cop_upper, fill=none, draw=none,forget plot] 
                             table{figsData/MF/MF_upper_traj_cop97.txt};

                     \definecolor{mycolor1}{RGB}{75,200,75}
                     \addplot[fill=mycolor1, fill opacity=0.5] fill between[of=cop_upper and cop_lower, ];  
                     \addlegendentry{Conformalized-DeepONet}
                     
                     \definecolor{mycolor2}{RGB}{0,255,255}
                     \addplot[fill=red, fill opacity=0.5] fill between[of=prob_upper and prob_lower, ];
                     \addlegendentry{Prob-DeepONet}                               
            \end{axis}
  \end{tikzpicture}
  \caption{Confidence intervals predicted by the Multi-fidelity conformalized Prob-DeepONet and the corresponding baseline model for a random high-fidelity test trajectory of the 1D jump function, given a miscoverage rate of $\alpha=0.05$. We summarize the average coverage rate for all approaches in Table \ref{table:multi-fidelity-coverage}. }
  \label{fig:multi-fidelity}
\end{figure}
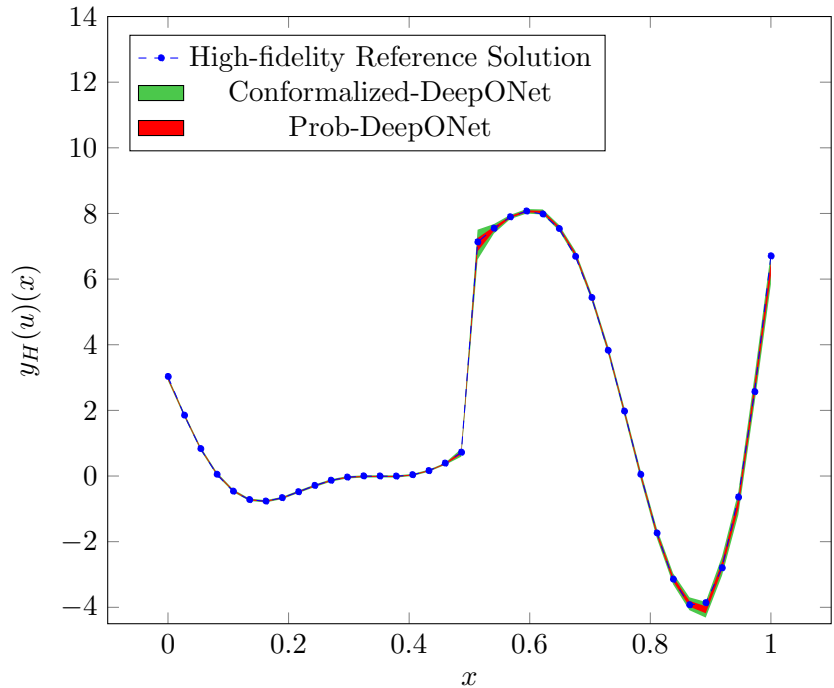

\textbf{Checking Coverage.} We tested the proposed Multi-Fidelity Conformalized Prob-DeepONet to create confidence intervals with coverage guarantees for a target miscoverage rate of $\alpha=0.05$. Figure~\ref{fig:multi-fidelity} shows the confidence intervals for a randomly chosen high-fidelity test trajectory. The Multi-fidelity Conformalized model provides a suitable confidence interval, showing increased uncertainty at the high-fidelity solution's jumps and turns. Additionally, Table~\ref{table:multi-fidelity-coverage} displays the average coverage of confidence intervals (targeting a miscoverage rate of $\alpha=0.05$), generated by both the Multi-fidelity conformalized and baseline Prob-DeepONet models for all $n_\text{traj}=100$ high-fidelity test trajectories. The results suggest satisfactory coverage, which can be enhanced by using more calibration examples.

\begin{table}[H]
\centering
\begin{tabular}{c   c   } 
\hline
\textbf{DeepONet model} &\textbf{Average coverage \% ($\alpha=0.05$) } \\
\hline
Multi-fidelity Conformalized Prob-DeepONet & \textbf{97.02}\% \\
Multi-fidelity Prob-DeepONet & 82.07\%  \\
\hline
\end{tabular}
\caption{The average coverage percentage for the predicted confidence intervals (with a given miscoverage rate of $\alpha=0.05$) calculated based on the $100$ high-fidelity test trajectories of the 1D jump function experiment.}
\label{table:multi-fidelity-coverage}
\end{table}


\textbf{On Our Future Work.} The findings discussed in this paper are undoubtedly promising and applicable in scenarios where a reliable surrogate model is required. Thus, in our future work, we plan to investigate the application of conformalized-DeepONets in real-world settings. This includes power systems~\cite{moya2023deeponet}, rib-optimization in fluid systems~\cite{sahin2024deep}, PDE-based topology optimization~\cite{lu2022multifidelity}, and forecasting extreme events. Additionally, we plan to investigate the use of conformal prediction in DeepONet settings where the properties of exchangeability and independent and identically distributed (i.i.d.) do not apply. For instance, in DeepONet extrapolation~\cite{zhu2023reliable}, DeepONet for non-autonomous systems~\cite{lin2023learning}, or Federated DeepONet~\cite{moya2022fed}. 

\section{Conclusion} \label{sec:conclusion}
In the paper, we presented methods to apply split conformal prediction for providing confidence intervals for Deep Operator Network (DeepONet) prediction. These intervals come with coverage guarantees and do not require distributional assumptions. Specifically, we used split conformal prediction to enhance our previously proposed Probabilistic and Bayesian DeepONets, enabling them to predict rigorous confidence intervals at a predetermined miscoverage rate. Moreover, we designed a novel extension of DeepONet, known as Quantile-DeepONet, which estimated conditional quantiles for DeepONet predictions and provided a more natural setting for applying conformal prediction. By combining split conformal prediction with Quantile-DeepONet, we successfully developed an effective and distribution-free methodology for constructing confidence intervals with guaranteed coverage.

\section{Ackownledgement}
G. Lin acknowledges the support of the National Science Foundation (DMS-2053746, DMS-2134209, ECCS-2328241, and OAC-2311848), and U.S. Department of Energy (DOE) Office of Science Advanced Scientific Computing Research program DE-SC0023161, the Uncertainty Quantification for Multifidelity Operator Learning (MOLUcQ) project (Project No. 81739) and DOE–Fusion Energy Science, under grant number: DE-SC0024583. 
L. Lu was supported by the U.S. Department of Energy [DE-SC0022953].

\bibliographystyle{unsrt}
\bibliography{references}
\end{document}

%% file: deepo.tex
\begin{tikzpicture}[scale = 1]
 \fill [green!10] (2, 4.5) rectangle (7, 7.6);
  \fill [blue!10] (2, 1.4) rectangle (7, 4.4);
    \fill [red!20] (-3.5, 2) rectangle (1, 7);

    \draw[ultra thick] [decorate,
    decoration = {calligraphic brace, mirror}] (-3.5, 1.8) --  (1, 1.8);
\node at (-1.2, 1.4) {\textcolor{red}{Inputs}};

\fill [magenta!10] (9, 4) rectangle (12, 5);
    \draw[ultra thick] [decorate,
    decoration = {calligraphic brace, mirror}] (9, 4) --  (12, 4);
\node at (10.6, 3.5) {\textcolor{magenta}{Outputs}};

\node[draw, text width=4cm] at (-1.2, 6) {$\hat{u} = [u(y_1), ..., u(y_N)]^\intercal$};

 \draw [-latex ](1,6) -- (2, 7);
 \node[draw, text width = 2.5cm] at (3.5, 7) {branch net $1$};
 \draw [-latex ](5, 7) -- (6, 7);
 \node[draw, text width = 0.5cm] at (6.5, 7) {$b_1$};
 
 \draw [-latex ](1,6) -- (2, 5);
  \node[draw, text width = 2.5cm] at (3.5, 5) {branch net $K$};
  \draw [-latex ](5, 5) -- (6, 5);
  \node[draw, text width = 0.5cm] at (6.5, 5) {$b_K$};
  
  \draw [-latex ](1,6) -- (2, 6);
 \node[draw, text width=2.5cm] at (3.5, 6) {branch net $j$};
 \draw [-latex ](5, 6) -- (6, 6);
\node[draw, text width = 0.5cm] at (6.5, 6) {$b_j$};

 \node[draw, text width = 2cm] at (-1, 3) {$x\in\mathbb{R}^d$};
  \draw [-latex ](1, 3) -- (2, 3);
 \node[draw, text width = 2.5cm] at (3.5, 3) {trunk net};
 
 \draw [-latex ](5, 3) -- (6, 4);
\node[draw, text width = 0.5cm] at (6.5, 4) {$t_1$};

  \draw [-latex ](5, 3) -- (6, 3);
\node[draw, text width = 0.5cm] at (6.5, 3) {$t_j$};

   \draw [-latex ](5, 3) -- (6, 2);
 \node[draw, text width = 0.5cm] at (6.5, 2) {$t_K$};

\draw [-latex ](7, 6) -- (9, 4.5);
  \draw [-latex ](7, 3) -- (9, 4.5);

  \node[draw, text width = 2.5cm] at (10.5, 4.5) {$\bigotimes\rightarrow G(u)(x)$};

\end{tikzpicture}